\documentclass[conference]{IEEEtran}
\ifCLASSINFOpdf
\else
\fi
\usepackage{graphicx}

\usepackage[T1] {fontenc}
\usepackage{textcomp}
\usepackage{lipsum}
\usepackage{eso-pic}

\makeatletter
\newcommand{\conf}[1]{%
\newsavebox{\headerbox}
\def\ps@IEEEtitlepagestyle{
\def\@oddhead{\parbox[b]{\dimexpr \textwidth-\wd\headerbox-\columnsep}{\centering \@IEEEheaderstyle #1}}%
\let\@evenhead\@empty
\def\@oddfoot{\mycopyrightnotice}%
\let\@evenfoot\@empty}}
\newcommand{\mycopyrightnotice}{\footnotesize 
979-8-3503-3015-1/23/\$31.00 $\copyright$2023 IEEE\hfill}
\makeatother

\conf{13th International Conference on Computer and Knowledge Engineering (ICCKE 2023), November 1-2, 2023, Ferdowsi University of Mashhad, Iran}

\begin{document}
%
\title{A Semi-supervised Fake News Detection using Sentiment Encoding and LSTM with Self-Attention}

\author{\IEEEauthorblockN{Pouya Shaeri}
\IEEEauthorblockA{Department of Computer\\and Data Sciences\\
	Shahid Beheshti University\\
	Tehran, Iran\\
	Email: p.shaeri@mail.sbu.ac.ir}
\and
\IEEEauthorblockN{Ali Katanforoush}
\IEEEauthorblockA{Department of Computer\\and Data Sciences\\
	Shahid Beheshti University\\
	Tehran, Iran\\
	Email: a\_katanforosh@sbu.ac.ir}
}


%


\maketitle

\begin{abstract}
Micro-blogs and cyber-space social networks are the main communication mediums to receive and share news nowadays. As a side effect, however, the networks can disseminate fake news that harms individuals and the society. Several methods have been developed to detect fake news, but the majority require large sets of manually labeled data to attain the application-level accuracy. Due to the strict privacy policies, the required data are often inaccessible or limited to some specific topics. On the other side, quite diverse and abundant unlabeled data on social media suggests that with a few labeled data, the problem of detecting fake news could be tackled via semi-supervised learning. Here, we propose a semi-supervised self-learning method in which a sentiment analysis is acquired by some state-of-the-art pretrained models. Our learning model is trained in a semi-supervised fashion and incorporates LSTM with self-attention layers. We benchmark our model on a dataset with 20,000 news content along with their feedback, which shows better performance in precision, recall, and measures compared to competitive methods in fake news detection.
\end{abstract}

%
\IEEEpeerreviewmaketitle

\section{Introduction}
The Internet and the cyber-space networks have increased the volume, speed, and variety of information available to the public, but, the potential for the spread of fake news has also increased, significantly. The platforms such as Facebook, Twitter, and YouTube, allow anyone to create and share contents with millions of users around the world without any editorial assessment or responsibility. Online platforms also allow users to filter and personalize their interested topics, preferences, biases, and to create an environment that adapted with their beliefs and opinions. Online platforms have been also exploited by malicious actors who create and spread fake news for various motives such as financial gain, ideological agenda, political influence or social disruption.

The challenge of detecting fake news has prompted researchers from different disciplines to develop methods and tools to automatically identify and disclose fake news. The algorithmic approaches for detecting fake news are generally classified into three categories: content-based methods, network-based methods, and hybrid methods.

Content-based methods focus on analyzing textual and visual features of online content to determine its authenticity. For example, content-based methods can use NLP techniques to detect linguistic signs of deception or sentiment analysis techniques to gauge the emotional tone of online content\cite{Alonso}. Content-based methods can also use computer vision techniques to detect manipulated images or videos\cite{Singh}.

Network-based methods focus on analyzing the social context and the spread pattern of online contents to determine their authenticity. For example, the node characteristics in complex network analysis and models for dynamic networks is used to classify users or news authenticity\cite{Zhou}.

Hybrid methods combine both content-based and network-based features to improve fake news detection. Hybrid methods like \cite{paka2021cross} use deep learning techniques to learn complex representations of online news content and social network characteristics of users or news sources with hybrid learning techniques.

Fake news detection is an active and interdisciplinary research area involving collaboration between computer scientists, social scientists, journalists, and policy makers. The problem of detecting fake news is an evolving and open-ended one that requires constant adaptation to new forms and sources of false and true information, as well as ethical considerations regarding privacy, freedom of expression, and social responsibilities.

Today, media and virtual networks have become a popular means for people to use and share news. Nevertheless, cyberspace also enables the widespread dissemination of fake news, i.e. news with intentionally false information, and creates significant negative effects on society. Fake news affects the individual as well as the whole society. Fake news can upset the balance of authenticity in the news ecosystem, persuading news audience to accept false or biased stories. For example, some people and organizations are recently spreading fake news on social media for financial and political gain. The effects of fake news sometimes cause irreparable events, such as news that causes physical violence or shootings in the United States. Also, the news of the death of a key figure of a country can have short-term destructive effects on the society until the detection of the fake news clears the confusion for the people. To alleviate this problem, research on fake news detection has recently received much attention. Despite several existing computational solutions for fake news detection, the lack of comprehensive and community-based fake news datasets has become one of the main obstacles for research. Not only the available datasets are scarce and unverifiable, but they are also limited to a specific topic or do not have many features required for the study, such as news content, image, listener feedback, and other features required for comprehensive news analysis\cite{shu2020fakenewsnet}.
Also, news on social media such as Twitter is generated at a high volume and speed. So that, very few can be labeled as fake or real news in a short period of time\cite{yang2019unsupervised}. In general, fake news detection could be categorized within the problems where we have little information about the authenticity or fakeness of a news item, and also in the case of a small number of news items, it is possible to state whether it is authentic or fake. Therefore, it is necessary to look at this issue as a semi-supervisory issue\cite{paka2021cross}.
Therefore, to overcome the problem of scarcity of datasets, we use a large-scale and evolved repository in terms of feature structure called FakeNewsNet, which includes comprehensive information with diverse features in news content, spatial information content social and time information and user page information. We provide a comprehensive description of FakeNewsNet, show exploratory analysis of the dataset from different perspectives, and discuss the advantages of FakeNewsNet for potential applications in the study of fake news in social media. Also, in order to deal with the second defect, we address the problem from perspective of a semi-supervised approach and design and implement a semi-supervised method on the introduced dataset.

In this paper, we intend to overcome these limitations by using the FakeNewsNet dataset and a novel self-learning semi-supervised deep learning network to detect fake news in social media. The advantages of our approach are as follows:

The FakeNewsNet dataset is a large-scale and comprehensive dataset containing news content, social context, and spatiotemporal information for studying fake news in social media. It covers two fact-finding websites, GossipCop and PolitiFact, and covers several topics, e.g., politics, entertainment, health, etc. It also provides the tweet IDs of the tweets sharing the news, which can be used to collect additional social media data.

The self-learning semi-supervised deep learning network is a new method that combines a trust network layer with an artificial neural network to detect fake news. The confidence network layer automatically returns and adds the correct results to help the neural network collect positive sample cases, thereby improving the accuracy of the neural network. Neural network uses graph structure of social media data to capture semantic and relational features of news articles and users \cite{li2021novel}. To improve this method, we use new layers such as LSTM with Self-Attention in the architecture of the artificial neural network. Also, to help accuracy and improve the detection of fake news, a sentiment analysis coding method has been used to improve this network.

Our self-learning semi-supervised deep learning network can handle the scarcity and imbalance of labeled data by using a small amount of labeled data and a large amount of unlabeled data. It can also adapt to new scenarios or domains by updating the trust network layer with new data \cite{li2021novel}.

Our self-learning semi-supervised deep learning network can combine the temporal dynamics and patterns of fake news dissemination using long short-term memory (LSTM) to model the sequence of tweets or news \cite{li2021novel}. Also, by using the attention mechanism, it can find out simple language structures among the words.

We expect that our approach can achieve higher performance and robustness in detecting fake news in social media.

\section{Proposed Algorithm}
In this section, we intend to improve the method described in \cite{li2021novel} by presenting a semi-supervised model employing a self-learning and pseudo-labeling algorithm and a novel transfer learning for sentiment analysis.\\
Due to the small number of labels in the dataset and the model-based labeling of the dataset, we cannot have absolute confidence in the labeling in semi-supervised methods. To improve accuracy and dependability, numerous techniques discard a portion of the data labeled by the model. The data loss standard in semi-supervised learning depends on the type of model and unsupervised network loss functions employed \cite{yang2021survey}. Pseudo-labeling, which assigns labels to unlabeled data based on the predictions of the \cite{liu2021swin} model, is a common technique for reducing data loss and enhancing reliable prediction.\\
However, noisy pseudo-labels can reduce model performance, so some methods employ confidence criteria to filter out low-confidence pseudo-labels or weight them differently in the loss function \cite{liu2021towards, wadawadagi2020self, tanlarge}.\\
In the proposed algorithm for the semi-supervised problem of detecting fake news, we assume that the input data has a small number of labels $(X,y) = \{D_l \cup D_u \}$, where $D_l$ are labeled samples in the training dataset of size $|L|$, then consider:
\begin{center}{$D_l^{(0)}=\{(X_1,y_1),(X_2,y_2),...,(X_l,y_l)\}$}
\end{center}

Also let $D_u$ denote the unlabeled samples in the test dataset of size $|U|$. Also consider:
\begin{center}
	$D_u = \{(X_{l+1},y_{l+1}),(X_{l+2},y_{l+2}),...,(X_{l+u},y_{l+u})\}$
\end{center}
Then the workflow of the self-learning semi-supervised deep learning algorithm can be described as follows:
\subsubsection{Initialization}
In the deep learning network, $D_l^{(0)}$ is used for learning and the learning process is carried out. Then in unsupervised deep network, pseudo-labels
$D'_u= \{(X_{l+1},\hat{y}_{l+1}), ((X_{l+2},\hat{y}_{l+2}), ..., (X_{l+u},\hat{y}_{l+u})\}$
are generated with the confidence value $\sigma$. If $\sigma_0$ is a threshold for limiting the selection of samples generated with high confidence such that the low confidence data in $D'_u$ is filtered out, then the pseudo-labeled selected set of $D'_u$ can be defined as follows with size $|P_0|$:
\begin{center}
	$D^{(0)}_{pseu}= \{(X_{l+i},\hat{y}_{l+i}), (X_{l+i+1},\hat{ y}_{l+i+1}),..., (X_{l+i+p},\hat{y}_{l+i+p})\}$\\
\end{center}
	
\subsubsection{Repetition loop}
Now we have the new training dataset as the union of previous training data and selected data labeled with high confidence value:
\begin{center}
	$D_l^{(1)}=|D^{(0)}_l \cup D^{(0)}_{pseu}|=\{(X_1,y_1),(X_1,y_1),\ldots,( X_l,y_l),\ldots,(X_{l+p},y_{l+p})\}$
\end{center}
This new dataset is used to retrain the deep learning network to generate a new pseudo-trusted label set $D^{(1)}_{pseu}$ of size $|P_1|$. This process is then repeated until the entire dataset has pseudo-labels with high confidence values.



\section*{Semi-supervised self-learning algorithm using pseudo-labeling}
In the initialization phase of the semi-supervised self-learning algorithm, we perform the steps outlined in \cite{li2021novel}. The division of the data into three sections, namely training data, evaluation data, and test data, is crucial and clarifies the situation significantly. Then, we divide the training data into k folds and assume that labels are only present in the first fold. If we assume k = 5, we can say that less than 20\% of the data we have is labeled and the problem is in semi-supervised mode.

\begin{figure}[htb!]
	\begin{center}		\includegraphics[width=0.5\textwidth]{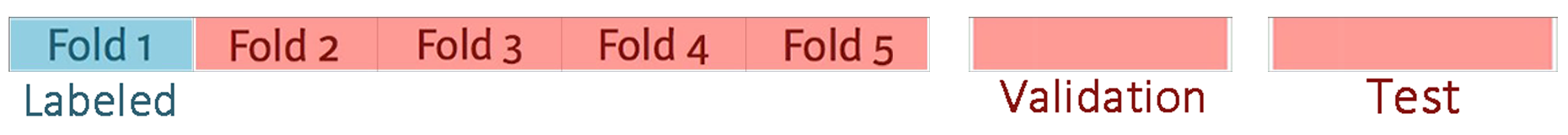}
		\vspace*{-0.7cm}
	\end{center}
	\caption[Folding the Dataset]{Folding the Dataset}
	\label{fig:DataFolding}
\end{figure}
Now, we fit the self-learning deep neural network model to the first fold's training data and evaluate using the fold's evaluation data. Now, using the trained network weights from the first fold, we predict the second fold's data.

\begin{figure}[htb!]
	\begin{center}		\includegraphics[width=0.5\textwidth]{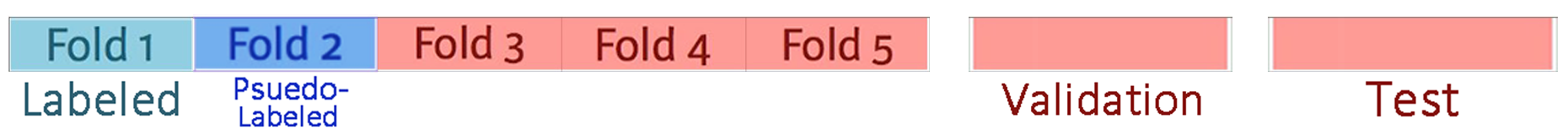}
		\vspace*{-0.7cm}
	\end{center}
	\caption[Fold 2 pseudo-labeled]{Fold 2 pseudo-labeled}
	\label{fig:DataFoldingFold2}
\end{figure}

Now by setting a confidence threshold
$0 < \sigma < 1$,
we retain the labels predicted by the neural network model from the second fold that are greater than the confidence threshold $\sigma$ or less than $1-\sigma$ (for example, 0.95). Now that we have the dataset from the first fold and pseudo-labeled data with a high degree of confidence, we call training on this dataset and evaluate it using the evaluation data. This procedure is repeated until the third, fourth, and fifth folds of pseudo-labeling are labeled securely.

\begin{figure}[htb!]
	\begin{center}		\includegraphics[width=0.5\textwidth]{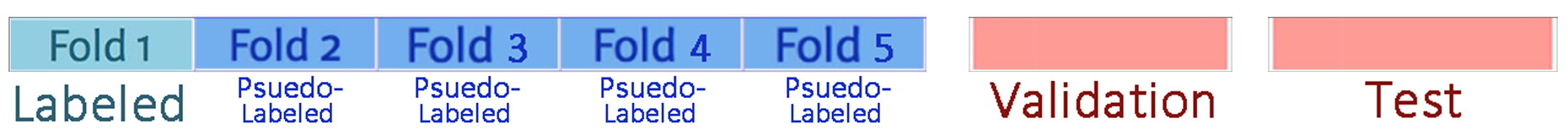}
		\vspace*{-0.7cm}
	\end{center}
	\caption[All folds pseudo-labeled]{All folds pseudo-labeled}
	\label{fig:DataFoldingAllFold}
\end{figure}

Now, by calling the proposed deep neural network, we train the data of the first through fifth folds and evaluate the network using test data.\\

This algorithm eliminates the ambiguity in the proposed method of the article \cite{li2021novel} that has been observed in numerous implementations. This ambiguity existed when collecting two sets of data from the pre-stage and post-stage fold, despite the evaluation and test data. In some existing implementations, for instance, in the first stage, after training the first fold, a portion of the first fold was selected for evaluation, and after pseudo-labeling, it combined randomly with the second fold for the input of the next stage of training, resulting in data leakage. To prevent this, we considered the validation dataset independently from the training data.

\section*{Proposed Model Architecture in Self-learning Semi-supervised Algorithm}
\begin{figure}[htb!]
	\begin{center}		\includegraphics[width=0.5\textwidth]{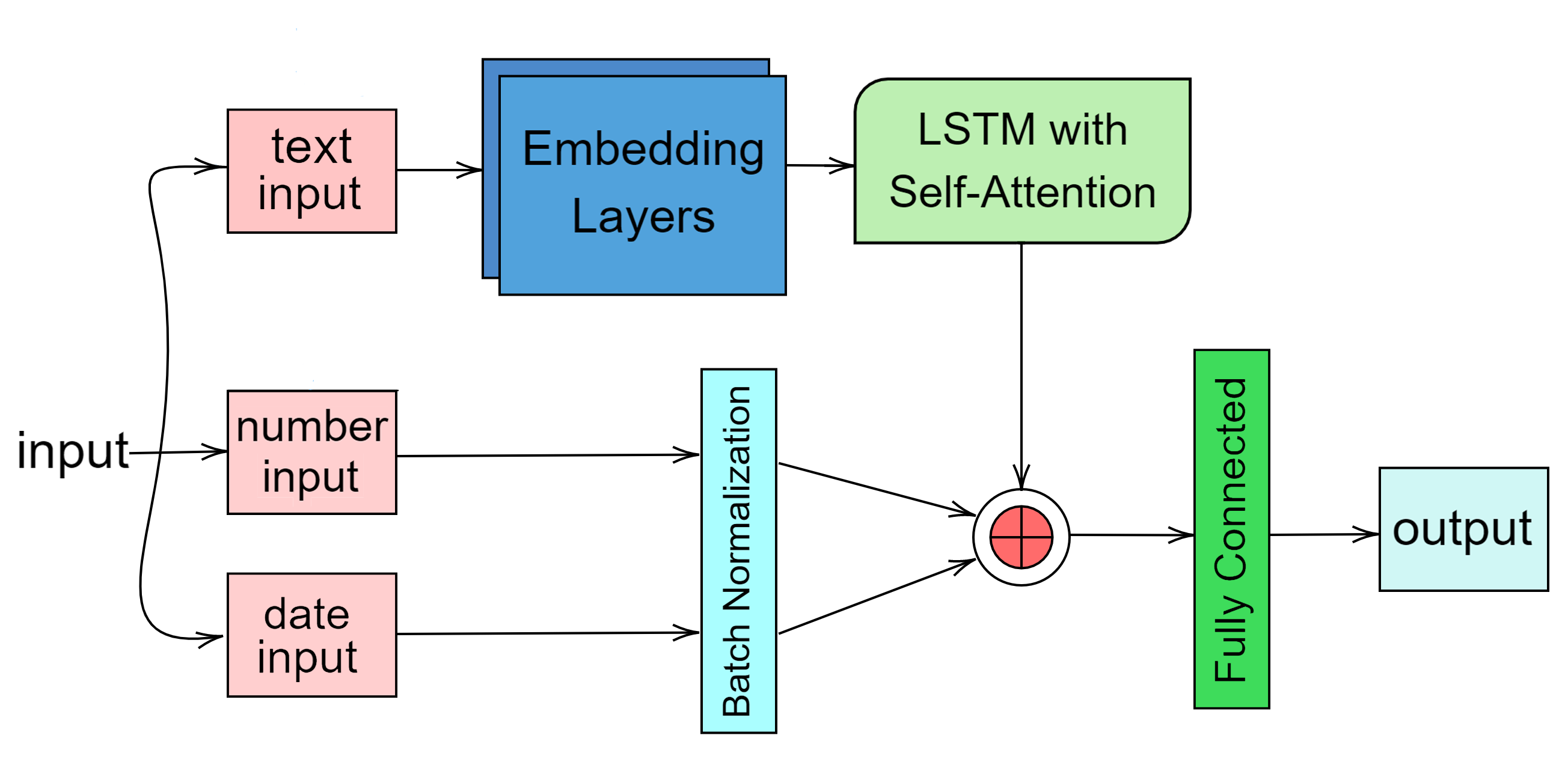}
		\vspace*{-0.7cm}
	\end{center}
	\caption[Proposed Deep Neural Network Architecture]{Proposed Deep Neural Network Architecture}
	\label{fig:SSArch}
\end{figure}

As stated previously, the majority of proposed deep neural networks for the problem of detecting fake news can be divided into three categories: content-based methods, Network-based methods, and hybrid methods.
The hybrid architecture of our proposed deep neural network for this classification is depicted in the Figure \ref{fig:SSArch}.

As can be seen from the Figure \ref{fig:SSArch}, this network receives three input features: textual data, numeric data and date. In this paper, we implemented the proposed artificial neural network architecture using Keras, which is a high-level neural network API that runs on top of TensorFlow, a popular open-source platform for machine learning. In this section, we discuss the architecture of different parts of the proposed model:

\subsection{Embedding Layer}
This layer of neural network and weighting receives news textual data, such as headline and text, news source, user's tweet text, other people's response text to that tweet, and some metadata information of users, such as the device/platform from which the tweet was posted, in the form of vectorization, padding, and tokenized. This layer converts positive integers to dense, fixed-size vectors. In fact, the text input layer is defined using the Input class from Keras to accept variable-length input sequences. The encoded input is then sent to an embedding layer, which transforms the input vector into a dense vector representation that is meaningful to the network.

\subsection{LSTM with Self-Attention Layer}
\begin{figure}[htb!]
	\begin{center}		\includegraphics[width=0.5\textwidth]{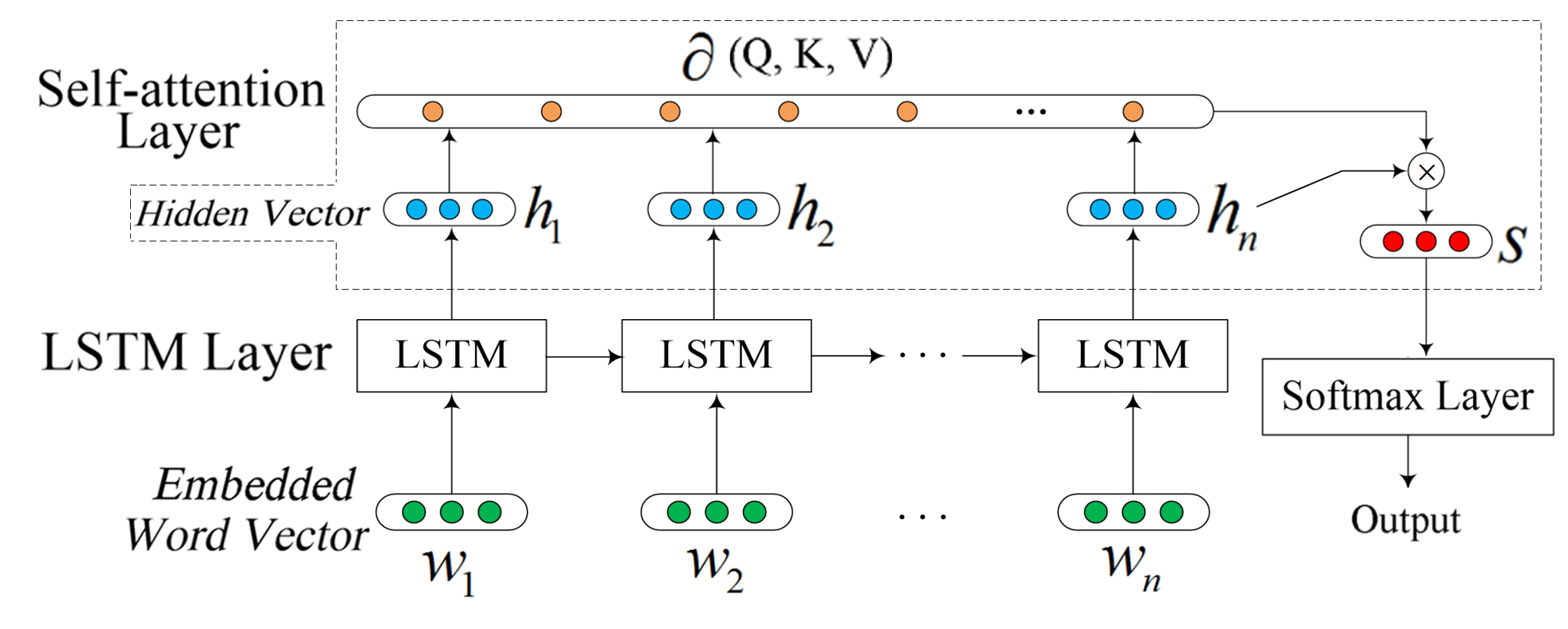}
		\vspace*{-0.7cm}
	\end{center}
	\caption[LSTM with Self-Attention]{LSTM with Self-Attention}
	\label{fig:LSTMwSA}
\end{figure}

Long Short-Term Memory (LSTM) with Self-Attention layer is a neural network architecture that combines the advantages of LSTM and Attention mechanisms. LSTM is a Recurrent Neural Network (RNN) that can discover long-term dependencies in sequential data, such as text, speech, or time series. Attention mechanisms are techniques that enable a network to concentrate on the most pertinent portions of input or output sequences and to learn how to align them. Self-attention is a subtype of attention in which the network attends to its own internal states as opposed to external inputs or outputs \cite{lstmwsa, vaswani2017attention}. 
A typical LSTM with Self-Attention layer includes an encoder LSTM that processes the input sequence and generates a sequence of hidden states, and a decoder LSTM that generates the output sequence based on the encoder states and its own previous states. Depending on the task \cite{LSTMWi}, the self-attention layer is applied to either the encoder states or the decoder states, or both \cite{LSTMWi}. The self-attention layer computes a weighted sum of all the states in a sequence, where the weights are learned based on the similarity between states (Figure \ref{fig:LSTMwSA}). In addition to capturing long-range dependencies and global information within a sequence, the self-attention layer can also reduce the dimension of the states.

In frameworks such as TensorFlow, Keras, and PyTorch, LSTM with Self-Attention layer can be implemented in various ways. One common approach is to use an existing attention layer, such as self attention or Attention in TensorFlow, and feed the same tensor twice as the query and value arguments, which results in self-attention. An alternative method is to write a custom layer that implements the self-attention formula, which involves computing the dot product between the states, applying a softmax function, and multiplying by the states once more. We implemented our LSTM with Self-Attention by defining the following states:

\begin{equation}
    u_t = tanh(W_w h_t + b_w)
\end{equation}
\begin{equation}
    \partial_t = \frac{exp(u_t^Tu_w)}{\sum_t exp(u_t^Tu_w)}
\end{equation}
\begin{equation}
    s=\sum_t \partial_t h_t
\end{equation}

\subsection{Fully Connected Dense Layers}
The Fully Connected Dense layer applies a linear transformation followed by an activation function to produce the output. The activation function then applies a nonlinear transformation to the previous layer's output. The output of the previous layer is then sent to a dense layer with ReLU activation. The final output of the model is generated by passing the output of the dense layer through a second dense layer with sigmoid activation.

	
	
	

\section*{Using Transfer Learning of Sentiment Analysis in Detecting the Fake News}
In this section, we will describe a transfer learning-based method for sentiment analysis and add it to the model proposed in this paper. Using labeled data from a source domain or task that is related to the target domain or task can help transfer learning overcome the problem of data scarcity. Transfer learning, for instance, can utilize labeled data from sentiment analysis of a text to detect fake news \cite{tlsa}. Text sentiment analysis is the process of identifying and extracting the opinions and emotions expressed in a text. The task of detecting and classifying news articles as real or fake based on their content and context is known as fake news detection.\\

The rationale for applying transfer learning from sentiment analysis of a text to the detection of fake news is that both tasks involve analyzing textual information and extracting useful features that can indicate the text's credibility and quality. For instance, the following features can be useful for both tasks:

\begin{itemize}
	\item Polarity and intensity of emotions expressed in the text
	\item Using emotional words, exaggeration or contradiction in the text
	\item The presence or absence of factual evidence, sources, or references in the text
	\item style, tone and readability of the text
\end{itemize}

A model can learn these features from a source domain or task and transfer them to a target domain or task using transfer learning. This can help improve model performance in the target domain or task by reducing the need for large quantities of labeled data and enhancing the model's generalization capability \cite{tlsa2}.

\begin{figure}[htb!]
	\begin{center}		\includegraphics[width=0.5\textwidth]{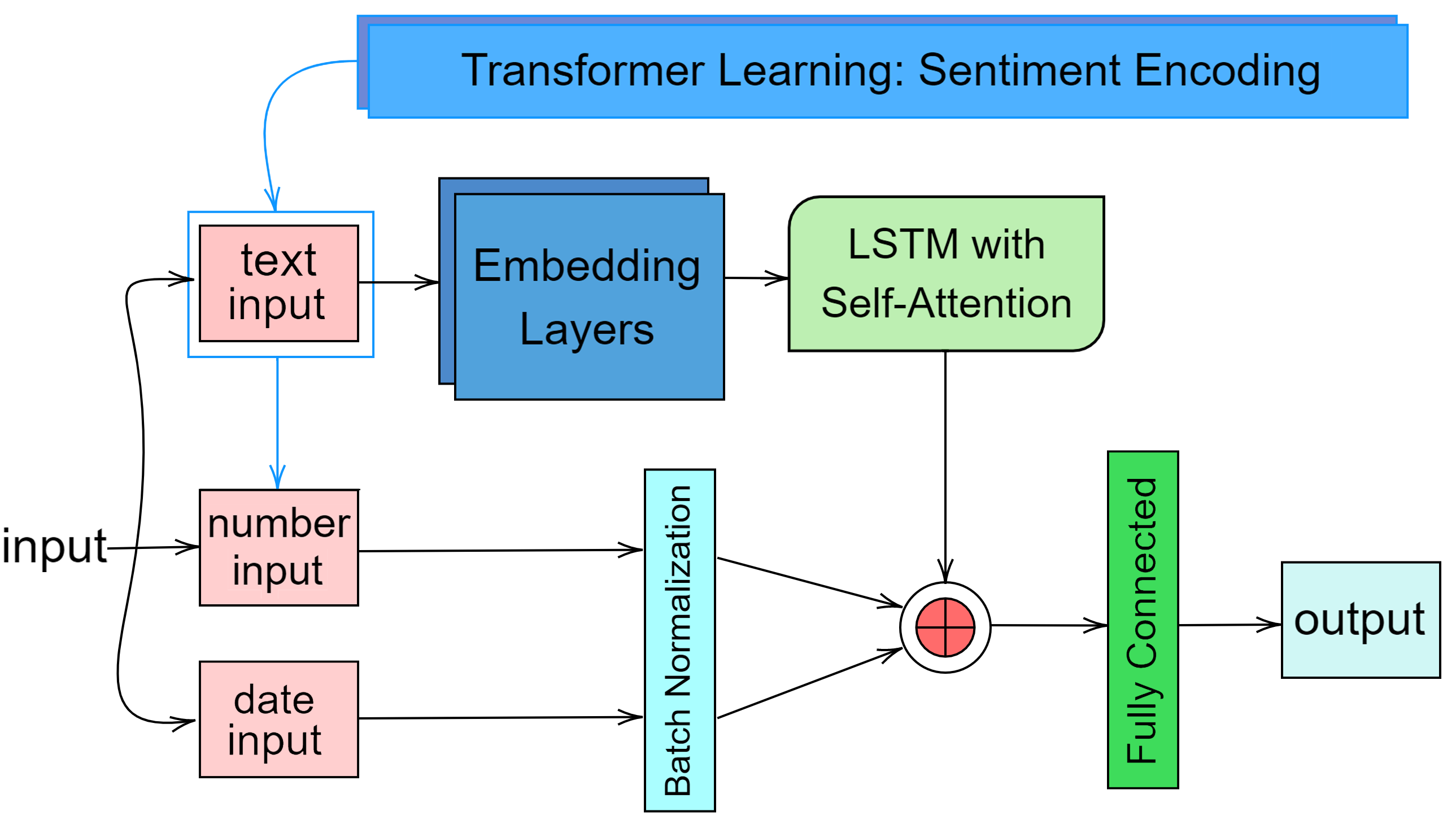}
		\vspace*{-0.7cm}
	\end{center}
	\caption[Proposed Deep Neural Network Architecture with Sentiment Encoding]{Proposed Deep Neural Network Architecture with Sentiment Encoding}
	\label{fig:SSArch2}
\end{figure}

In this paper, we intend to implement the prediction of a transfer learning based on sentiment analysis on two columns of news text and tweet text of FakeNewsNet data with a coding method. We have used two pre-trained pipelines from RoBERTa (Robustly Optimized BERT Pre-training Approach) specifically for the news text and the tweets \cite{roberta}. Then, we encode the two columns to the nominal features of sentiment analysis in the dataset (Figure \ref{fig:SSArch2}). This process is called Sentiment Encoding.

In fact, a hidden pattern of sentiment in a text according to the news text and tweet text helps the deep neural network in the semi-supervised self-learning model to label and identify fake news using text sentiment analysis.

\section{Data Description}
This paper's proposed model receives our crawled and feature extracted FakeNewsNet benchmark dataset, which includes features such as textual news content, a large number of tweets, responses, and sentiment.
Due to Twitter's privacy policies, this dataset initially consists only of labels, news headlines, news source, and tweet IDs. All news information including text, date, photo and video of news and metadata of Twitter users, as well as the text and tweet associated with each news, extracted by the Crawler extractor located on the Github page of this dataset's creators \cite{shu2020fakenewsnet} with the preparation of this dataset. We were also able to crawl and extract this dataset using the Twitter API. However, it should be noted that after two years from the release date of this dataset, there are some issues with the news and tweets included in this dataset.
The following issues can be mentioned as part of these:

\begin{itemize}
	\item
	Twitter has suspended many users from this virtual space in the last two years. Most of those who have been suspended are those who have been inactive in this social space for some time. Therefore, the tweets and metadata of these people are not available.
	\item
	Many may have disabled or made their account private.
	\item
	Many people may have deleted tweets related to the news for which we have the tweet ID.
	\item
	Many news have been deleted from the relevant news website page.
\end{itemize}
Despite these obstacles, we were able to set up and run a crawler to crawl this dataset and the FakeNewsNet dataset by taking into account and overcoming the limitations. After cleaning up the data in the feature columns, we selected key variables to measure and compare for our proposed model. These include the title, text, source, and date of the news and the text of the tweet related to the news, the number of retweets, the user name of the user who posted the tweet, the response texts to the corresponding tweet, the date of the tweet, and from the user page information, the number of tweets and the number of followers, number of followings, number of likes, and the user's registration date on Twitter, as well as the means by which the tweet was sent.

\section{Expermiental Results}
In this section, we will first examine the outcomes of the proposed model's evaluation. The results obtained in the article introducing this dataset will then be examined. Then, we examine the outcomes of semi-supervised models with various architectures on this dataset and compare them to the outcomes of the proposed method. The evaluations were carried out on a device with 13GB of RAM and an Intel Xeon CPU processor with a base clock of 2.2 GHz and an  NVIDIA P100 GPU with 16 GB VRAM.

\subsection{Evaluation of the Proposed Model}
In this paper, to evaluate the semi-supervised self-learning model, the fold-by-fold results, which are all on data outside of the training data even when they are not in the fold, are given in Table \ref{tab:SLSS}.

\begin{table}[!ht]
	\centering
	\caption[Evaluation of the Proposed Model]{Evaluation of the Proposed Model}
	\label{tab:SLSS}
		\begin{tabular}{c c c c c}
			\hline
			FOLD & Accuracy & Precision & Recall & F1-Score \\
			\hline
			Fold1-Val & 0.8410 & 0.8533 & 0.8503 & 0.8513 \\
			
			Fold+2-Val & 0.8540 & 0.8601 & 0.8630 & 0.8603 \\
			
			Fold+3-Val & 0.8520 & 0.8655 & 0.8792 & 0.8772 \\
			
			Fold+4-Val & 0.8680 & 0.8714 & 0.8799 & 0.8753 \\
			
			Fold+5-Test & 0.8619 & 0.8736 & 0.8667 & 0.8701 \\
			
			\hline	
		\end{tabular}
\end{table}

\subsection{Comparison of the Proposed Model with Previous Works}
Now, we apply fundamental machine learning methods such as Logistic Regression, Naive Bayes, SVM, and Random Forrest to the extracted and preprocessed FakeNewsNet dataset of this paper and report the results of this observational method in Table \ref{tab:FNNResults}. Then, we'll compare it to the outcomes of the proposed method applied to the crawled data.

\begin{table}[!ht]
	\centering
	\caption[Results of the basic machine learning methods]{Results of the basic machine learning methods}
	\label{tab:FNNResults}
		\begin{tabular}{c c c c c}
			\hline
			Method & Accuracy & F1-Score \\
			\hline
			
			Logistic regression & 0.740 & 0.751 \\
			
			Naive Bayes & 0.720 & 0.715 \\
			
			SVM & 0.762 & 0.753 \\
			
			Random Forrest & 0.732 & 0.750 \\
			
			\hline
		\end{tabular}
\end{table}
Notably, the results \ref{tab:FNNResults} of the methods used in this article were obtained using a supervised approach on this dataset. Now, we'd like to examine the outcomes of the presented semi-supervised models with various architectures on this dataset and compare them to the outcomes of the proposed method. The outcomes of the methods described in the article \cite{li2021novel} are listed in Table \ref{tab:RNN}.
\begin{table}[!ht]
	\centering
	\caption[The results of the methods of the \cite{li2021novel}]{The results of the methods of the \cite{li2021novel}}
	\label{tab:RNN}
		\begin{tabular}{c c c c c}
			\hline
			Method & Precision & Recall & F1-Score \\
			\hline
			
			CNN & 0.71 & 0.69 & 0.79 \\
			
			BI-LSTM & 0.73 & 0.71 & 0.74 \\
			
			LSTM/RNN & 0.86 & 0.83  & 0.81 \\
			
			\hline	
		\end{tabular}
\end{table}
We reimplemented the LSTM and implemented the proposed LSTM with Self-Attention method in a semi-supervised self-learning algorithm in this paper. The comparison of the results is presented in Table \ref{tab:OurMethod}.

\begin{table}[!ht]
	\centering
	\caption[Results of our method]{Results of our method}
	\label{tab:OurMethod}
		\begin{tabular}{c c c c c}
			\hline
			Method & Precision & Recall & F1-Score \\
			\hline
			
			LSTM & 0.84 & 0.83 & 0.84 \\
			
			LSTM with Self Attention & \textbf{0.87} & \textbf{0.86} & \textbf{0.87}\\
			\hline	
		\end{tabular}
\end{table}
By examining the results in Table \ref{tab:OurMethod}, we have determined that the proposed method yields superior outcomes compared to the previous methods. This is also due to the fact that the features extracted from this dataset, such as the user's Twitter registration date and the device with which the tweet was uploaded, as well as the six coded columns of the sentiment analysis of the news text and tweet and pre-processing with the most up-to-date methods, aid in solving the problem to a significant degree.

\section{Conclusion}
In this paper, we present a self-learning semi-supervised learning method for fake news detection using transfer learning of text sentiment analysis. The problem of fake news detection in the real world is inherently a semi-supervised problem. Because, in reality, we never have a labeled dataset with numerous features and diverse topics. This method pseudo-labels the unlabeled data with a high confidence threshold in a semi-supervised algorithm. We also tried to use up-to-date and more efficient layers, such as the LSTM with Self-Attention layer, in the self-learning architecture to increase the accuracy and efficiency of the model.

To obtain the dataset, we faced some limitations that we solved and finally worked on a comprehensive dataset in terms of feature structure and number. Then we preprocessed this dataset using modern methods and feature extraction and implemented a semi-supervised self-learning algorithm using deep neural networks on it as a semi-supervised problem. In addition to overcoming the mentioned limitations, by using transfer learning on sentiment analysis and adding it to the model, the performance and accuracy of the model increased significantly, which is evident in the results obtained in the performance metrics of the proposed model.

\subsection*{Future Work}
There are different paths for future work that hold significant value for academic exploration. In the current work, we considered that in the semi-supervised algorithm, only the training data were folded, and the test data were fixed and only measured at the final stage. Also, the confidence threshold variable is assumed to be constant. But we can also have such folding or cumulative folding for test data at each test stage and update the confidence threshold adaptively at each stage. Also, in the sentiment encoding section, we can involve more models to have a more unique encoding with a wider range of negative, neutral, and positive sentiments within the encoded vector. As another direction for future work, it is recommended to build upon the previous work that discontinued data crawling three years ago. We can introduce a complete and diverse dataset by using the crawler, more websites than the two used in previous work, and newer news and adding more diverse topics such as sports news, financial news, etc.






%

\vspace{0.5cm}

\bibliographystyle{IEEEtran}
\bibliography{references}

\end{document}